\documentclass{article}

\usepackage{PRIMEarxiv}

\usepackage[utf8]{inputenc} 
\usepackage[T1]{fontenc}    
\usepackage{hyperref}       
\usepackage{url}            
\usepackage{booktabs}       
\usepackage{amsfonts}       
\usepackage{nicefrac}       
\usepackage{microtype}      
\usepackage{lipsum}
\usepackage{fancyhdr}       
\usepackage{graphicx}       
\graphicspath{{media/}}     
\usepackage{amsmath}
\usepackage{caption}
\usepackage{subcaption}
\usepackage{multirow}

\title{Scalable Graph Self-Supervised Learning
}

\author{
Ali Saheb Pasand$^{*}$ \\
University of Waterloo 
\And
Reza Moravej$^{*}$ \\
University of Toronto 
\And
Mahdi Biparva$^{\dagger}$ \\
Huawei Noah's Ark Lab
\And
Raika Karimi \\
Huawei Noah's Ark Lab
\And
Ali Ghodsi \\
University of Waterloo \\ \\
$*$ done during internship $\dagger$ correspondence to \texttt{mahdi.biparva@huawei.com}
}

\begin{document}
\maketitle

\begin{abstract}
In regularization Self-Supervised Learning (SSL) methods for graphs, computational complexity increases with the number of nodes in graphs and embedding dimensions. To mitigate the scalability of non-contrastive graph SSL, we propose a novel approach to reduce the cost of computing the covariance matrix for the pre-training loss function with volume-maximization terms. Our work focuses on reducing the cost associated with the loss computation via graph node or dimension sampling. We provide theoretical insight into why dimension sampling would result in accurate loss computations and support it with mathematical derivation of the novel approach. We develop our experimental setup on the node-level graph prediction tasks, where SSL pre-training has shown to be difficult due to the large size of real world graphs. Our experiments demonstrate that the cost associated with the loss computation can be reduced via node or dimension sampling without lowering the downstream performance. Our results demonstrate that sampling mostly results in improved downstream performance. Ablation studies and experimental analysis are provided to untangle the role of the different factors in the experimental setup. 

\end{abstract}

\section{Introduction}







The problem of learning on graphs is of practical importance in application domains such as large-scale social networks, recommender systems, knowledge graphs, and biological/biomedical networks \cite{hamilton2017inductive,hu2020open,hu2021ogb,huang2023temporal}. However, in many such application areas, access to downstream labels may not be possible, especially when dealing with graphs of enormous sizes. Recent research has investigated self-supervised pre-training on graphs \cite{liu2022graph}. However, recent work on the graph domain has only been restricted to moderate-size graphs due to the high computational cost of pre-training graph encoding models on larger graphs. To this end, having efficient pre-training methods supported by solid theoretical bases to scale up the self-supervised learning methods 
to large graphs is crucial. 

The spectrum of scalable approaches for self-supervised pre-training of graphs is generally too narrow for graphs of significant sizes. Reconstruction-based methods generally formulate 
the pretext task based on graph reconstruction. However, the cost incurred by the reconstructing decoder is substantial for large graphs. A viable alternative is 
the joint-embedding methods, which learn invariance to augmented versions of the same input. Though more efficient than the reconstruction based methods, the cost of training a joint-embedding model on graphs is still significant. 





The joint-embedding 
non-contrastive Self-Supervised Learning (SSL)
methods relies on regularizing the invariance loss terms applied on the positive examples in order to prevent dimensional collapse \cite{NEURIPS2022_aa56c745,garrido2023rankme,DBLP:journals/corr/abs-2105-00470}. 
Compared to the contrastive self-supervised setting, the computational cost (in terms of both the time and space complexity) of such SSL methods 
stem from two operations: 
$(i)$ joint-embedding approaches take advantage of augmented views of the same input. Thus, multiple passes to the encoder is required across different such methods. This makes the space and time complexity of the encoder computation 2 to 3 times 
the encoder computation in the supervised domain. $(ii)$ there is often a high time and memory cost associated with computing the SSL loss functions in non-contrastive methods which is imposed by the calculation of the regularization terms. 
On the other hand, it is also hypothesized that the embedding space should be large enough to cover the semantic space of the downstream task \cite{haochen2021provable,bardes2022vicreg,zbontar2021barlow}. 
The computational burden is more challenging to overcome when the condition of projecting the input samples to a sufficiently high-dimensional embedding space on which the pre-training loss function is minimized, is necessary to satisfy.
The computational cost of calculating the covariance matrix for the regularization terms in volume maximization approaches such as VICReg \cite{bardes2022vicreg} and Barlow Twins \cite{zbontar2021barlow} significantly increases as the embedding space expands into larger dimensions. This even makes dimensional collapse of a serious concern that should be properly addressed.

In this work, we investigate methods to resolve the two aforementioned computational bottlenecks to efficiently scale self-supervised learning on graphs with various scales. We turn our attention to Information (Volume) Maximization methods \cite{bardes2022vicreg,zbontar2021barlow,bielak2022graph}, which do not require architectural tricks to prevent dimensional collapse in comparison to distillation methods \cite{grill2020bootstrap,thakoor2022largescale}, and require fewer samples in mini-batches for calculating the loss in comparison to contrastive methods \cite{DBLP:conf/icml/ChenK0H20,veličković2018deep,zhu2020deep}. We aim to reduce the extra cost associated with computing the regularization terms in the loss in Information Maximization approaches via sampling nodes and features dimensions. We interestingly find that the performance of models trained with sampling surpasses the performance of models trained with no sampling. 

In summary, our contributions can be summarized as follows. We provide analysis on the computational complexity of the covariance matrix in Sec. \ref{sec:background}. In Sec \ref{3}, we introduce our novel approach by focusing on reducing the computational complexity via node sampling. We shed light on why nodes sampling would work in the self-supervised domain, and to what extend can sampling be applied without sacrificing accuracy. We provide empirical evidence across a variety of graph benchmark datasets. 
In section \ref{sec:dimsam}, we present the formulation behind dimension sampling and how it is related to the Nystrom approximation for scalable manifold learning.
We show how a simple uniform sampling can result into lower computational cost while maintaining the performance of the downstream task.

\section{Related Work}

\subsection{Self-Supervised Representation Learning}
The recent line of work on SSL has introduced a variety of approaches for unsupervised pre-training. Contrastive methods minimize the similarity of the embeddings of different samples while maximizing the similarity of different views of the same sample \cite{DBLP:conf/icml/ChenK0H20}. Clustering approaches perform contrastive learning at the level of clusters rather than individual samples. clustering-based methods group samples into clusters based on some similarity measure \cite{10.5555/3495724.3496555,HuangDGZ19,DBLP:conf/nips/CaronMMGBJ20}. Distillation methods \cite{DBLP:journals/corr/abs-2006-07733,chen2020simsiam} take a different route and prevent collapse via architectural tricks. Distillation approaches propose an online network along with a target network, where the target network is updated with an exponential moving average of the online network weights. The line of work by \cite{bardes2022vicreg,DBLP:journals/corr/abs-2103-03230,ermolov2021whitening} propose information maximization approaches which maximize the information content of the embeddings. Such methods produce embedding variables that are decorrelated from each other, thereby preventing an informational collapse in which the variables carry redundant information. The downside of whitening methods such as VICReg \cite{bardes2022vicreg} and Barlow Twins \cite{zbontar2021barlow} is their dependence on a large projection head.

\subsection{Efficient Graph Training}
Prior work on efficient graph training have focused on the supervised setting. Real-world graphs often consists of millions on nodes and edges, which makes training GCN models on them infeasible. As such, sampling based methods have been proposed. Prior work have proposed layer sampling \cite{10.5555/3327345.3327367,10.5555/3495724.3496944}, subgraph sampling \cite{10.1145/3292500.3330925,Zeng2020GraphSAINT}, and neighbour sampling \cite{hamilton2017inductive,chen2018fastgcn}. A different line of research has proposed leveraging multiple GPUs to enable full-graph training. The focus of our work has been on the whitening loss term in the self-supervised setting rather than enhancing the efficiency of 
the graph encoder. We note that the efficient encoding techniques can be used in parallel to our direction in order to make the end-to-end training of the model more efficient.

\section{Background}
\label{sec:background}
\subsection{Problem Formulation}

The pre-training stage is where the SSL model is trained on unlabelled data to learn representations which can later be used for a downstream task. Consider a graph $\mathcal{G}$ defined by a tuple $(V, A, X)$, where $V$ is the set of nodes in the graph, $A \in \{0,1\}^{N \times N}$ is the adjacency matrix representing the edges in the graphs, and $X \in \mathbb{R}^{N \times F}$ are the node features. We apply two stochastic graph augmentations, $\mathcal{T}^1$ and $\mathcal{T}^2$, to $\mathcal{G}$, obtaining two views: $\mathcal{G}^1 = (V^1, A^1, X^1)$ and $\mathcal{G}^2 = (V^2, A^2, X^2)$. The SSL model consists of a graph encoder $f_\theta : \mathbb{R}^{F} \mapsto \mathbb{R}^{S}$ and a projector/expander $g_\phi: \mathbb{R}^{S} \mapsto \mathbb{R}^{D}$. 
We denote the representations $H \in \mathbb{R}^{N \times S}$ as $H = f(\mathcal{G})$ and the embeddings as $Z \in \mathbb{R}^{N \times D}$, where $Z = g_\phi(f_\theta(\mathcal{G}))$. In general, Graph SSL approaches minimize a loss function of the form: 
\begin{equation}
 \mathcal{L(\theta, \phi)} = \mathbb{E}_{v_1 \sim V_1, v_2 \sim V_2} sim(z_1, z_2)
\end{equation}
where $sim$ is a similarity function such as cosine or euclidean similarity.
After pre-training is completed, the projector $g_\phi$ is often thrown away. The representations learned by the trained encoder $f_{\theta}$ can be used on any graph downstream task, such as node classification or edge prediction. 
We turn our attention to VICReg, which explicitly regularizes the embedding space to prevent collapse. The VICReg loss consists of three terms as follows:

\textbf{Invariance:} a term to enforce similarity between different positive views.
\begin{equation}
  I(Z^1, Z^2) = \frac{1}{N} \sum_{i = 1}^{N} \| z_i^1 - z_i^2 \|
\end{equation}

\textbf{Variance:} a term which maximizes the volume of the distribution of the samples in the embedding space. The variance is written as a hinge loss to maintain the standard deviation (over a batch) of each variable of the embedding above a given threshold.
\begin{equation}
V(Z) = \frac{1}{d} \sum_{i=1}^d max(0, 1-f(z^j, \epsilon)),\\\quad\quad
\end{equation}
\begin{equation}
f(z, \epsilon) = \sqrt{(var(z) + \epsilon)}
\end{equation}

  Where $z^i$ is the vector consisting of each value at dimension $i$ in vectors in $Z$.

\textbf{Covariance:} a term to prevent informational collapse, where the variables would be highly correlated. The covariance term decorrelates the variables of each embedding.
\begin{equation}
  C(Z) = \frac{1}{d}\sum_{j\neq j} [Cov(Z)_{i,j}^2]
\end{equation}
  Where $Cov(Z)$ is the covariance matrix of $Z$.


  


The computational complexity of the invariance and variance is $O(ND)$, 
where $N$ is the number of nodes in the graph and $D$ is the embedding dimension coming out of the expander module in VICReg. However, the complexity of computing the covariance matrix is $O(ND^2)$. Since $D$ often is quite large and quadratic, computing this term could incur significant computational and memory costs. Additionally, $N$ is several orders of magnitude larger than $D$;
hence, controlling it can significantly impact the efficiency as well.

\subsection{Computational Complexity}
\label{sec3}

To motivate our work, we provide a brief description on the time and space complexities of different graph SSL models in Table \ref{tab:comp}. For a graph with $N$ nodes and $E$ edges, we consider a graph encoder which computes embeddings in time and space $O(N + E)$. This property is satisfied by most GNN encoders such as convolutional \cite{kipf2017semisupervised}, attentional \cite{veličković2018graph}, and message passing \cite{10.5555/3305381.3305512} networks. Distillation methods (such as BGRL \cite{thakoor2022largescale}) perform four encoder computations per update step (twice for the target and online encoders, and twice for each augmentation), contrastive (e.g. GRACE) and information Maximization methods (e.g. VICReg) both perform two encoder computations (once for each augmentation). All methods perform a node-level projection step. While BGRL and GRACE backpropagate the gradients twice (once for each augmentation), VICReg with identical architectures and shared weights requires one backpropagation. Assuming the backward pass to be approximately as costly as the forward pass, the time and space complexities of different models can be summarized in Table \ref{tab:comp} (We ignore the cost for computation of the augmentations):

\begin{table}[t]
\centering
\label{compTable}
\resizebox{0.6\columnwidth}{!}{%
\begin{tabular}{c|ccc}
\hline 
Method                        & Encoder                           & Prediction & SSL Loss                \\ \hline \midrule
BGRL                         & $O(6. (N + E))$                        & $O(4 . N)$      & $O(N)$                       \\
Grace                          & $O(3. (N + E))$                        & $O(4. N)$       & $O(N^2)$    \\
VICReg (non-Siamese) & $O(4. (N + E))$ & $O(4. N)$       & $O(N.D^2)$*  \\
VICReg (Siamese)   & $O(3. (N + E))$   & $O(4. N)$       & $O(N.D^2)$*
\end{tabular}
}
\caption{Computational complexities of several SSL approaches. N and E denote the number of nodes and edges in the input graph. D is output dimension of the predictor/expander module. Constants values represents the number of passes of a module. * The space complexity for the VICReg loss is $O(D^2)$}
\label{tab:comp}
\end{table}


We note that while distillation methods have the most efficient loss computation, they require 6 passes to the encoder due to the architectural tricks. The contrastive loss has a $O(N^2)$ complexity, which can be expensive for large $N$. While neighborhood sub-graph sampling can bring down this cost, it may result in loosing useful information in extremely large graphs or graphs with important topological information. As such, we turn our attention to information maximization approaches, which are the most computationally feasible due to fewer encoding passes per update step and a $O(ND^2)$ loss. State-of-the-Art information maximization approaches such as VICReg depend on a large projection head to compute the final loss in high dimension. We demonstrate dimension sampling can be used to alleviate the loss calculation in high-dimensional embedding spaces.


\section{Node Sampling}
\label{3}

We first being by investigate sampling nodes in order to reduce the complexity of computing and storing the covariance matrix. We note that unlike node sampling techniques which sample nodes prior to feeding the input to the encoder, we perform sampling at the final layer of the SSL head (expander or predictor depending on the SSL method). This ensures that we don't get a high variance of node feature estimations, which could hurt the model performance \cite{10.1145/3394486.3403192}. To randomly distribute the contribution of nodes to the loss calculation, we repeat sampling at each epoch. We briefly investigate two methods for node selection. We leave a more rigorous empirical investigation on node sampling using node pooling \cite{grattarola2022understanding} based on SSL requirements to future work. We consider uniform sampling as a simple baseline for node selection, and introduce Ricci sampling as more sophisticated alternative. In section \ref{sec5}, we test the impact of these node sampling techniques via computing the SSL loss on the selected subset of nodes.

\textbf{Uniform Sampling:} We first consider the simplest sampling procedure where a ratio of all the nodes in the input graph are chosen using a binary mask sampled from a uniform distribution (the probability of a node being selected is $\frac{1}{N}$). We aim to test our hypothesis that how node sampling impacts representation learning in the pre-training stage in terms of the performance of the model on the downstream tasks.

\textbf{Ricci Sampling:} If the embeddings of a node is similar to the embeddings of the augmented views of the node, then the node does not carry information about the data augmentation; hence, it is better to be dropped for loss minimization. This is due to the fact that the invariance loss would be applied to two similar vector representations, which has little impact on the learning process. We devise a simple method based on Ricci curvature which satisfies the above criteria. Previous work generalizing the notion of Ricci curvature (of Riemannian manifolds) to graphs has shown that the a graph with large/small curvature corresponds to the edge which is more/less connected than a grid \cite{topping2022understanding}. We use the method proposed by \cite{Topping2021UnderstandingOA} to calculate Ricci curvature of each edge. We denote the total Ricci flow of node $n_i$ as the summation of Forman Ricci Flows of edges connected to the node. According to \cite{Topping2021UnderstandingOA} edges with negative Ricci flow are subject to be over-squashed and hinder the flow of information through the message passing in GNN. Thus, if a node is connected to many edges with negative Ricci flow, the change in embedding and structure of different parts of graph will not be propagated to that node (a node with negative Ricci flow would be overflowed with information). Consequently, nodes with negative (or small positive) Ricci flows won't have their embedding change significantly after augmentations. Thus, it is permissible to sample nodes with positive Ricci flows more frequently. We sample node $n_i$ with probability $Pi(n_i) = \frac{Ricci(n_i) - min_{j} Ricci(n_j)}{\sum_{j = 1}{N} Ricci (n_j)}$ (normalize Ricci to get a valid probability distribution, shift to right in order to have minimum Ricci as zero and normalizing to sum to one.)

\section{Dimension Sampling}
\label{sec:dimsam}
\begin{figure}[!h]
  \centering
    \includegraphics[width=0.20\textwidth]{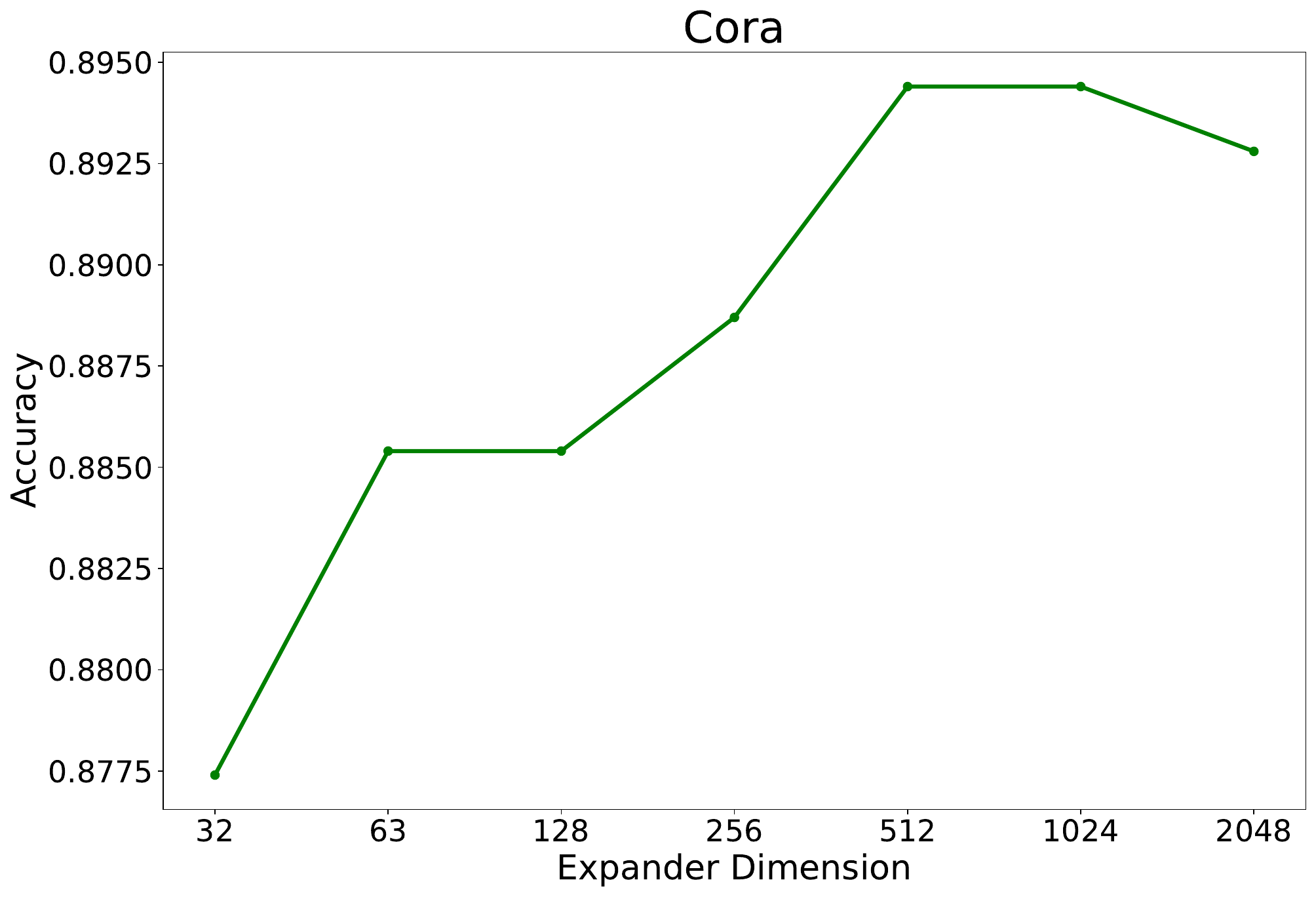}
    \includegraphics[width=0.20\textwidth]{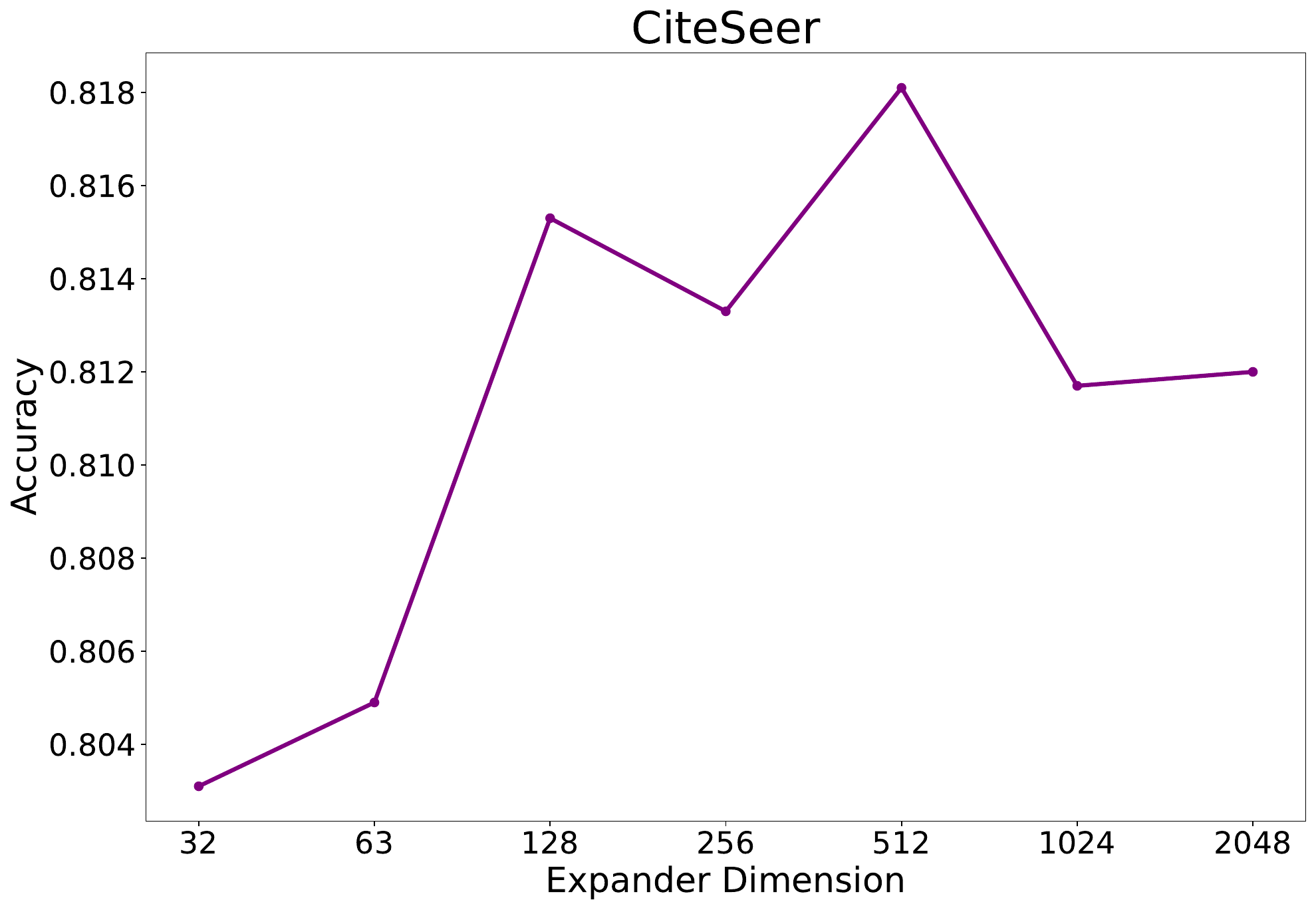}
    \includegraphics[width=0.20\textwidth]{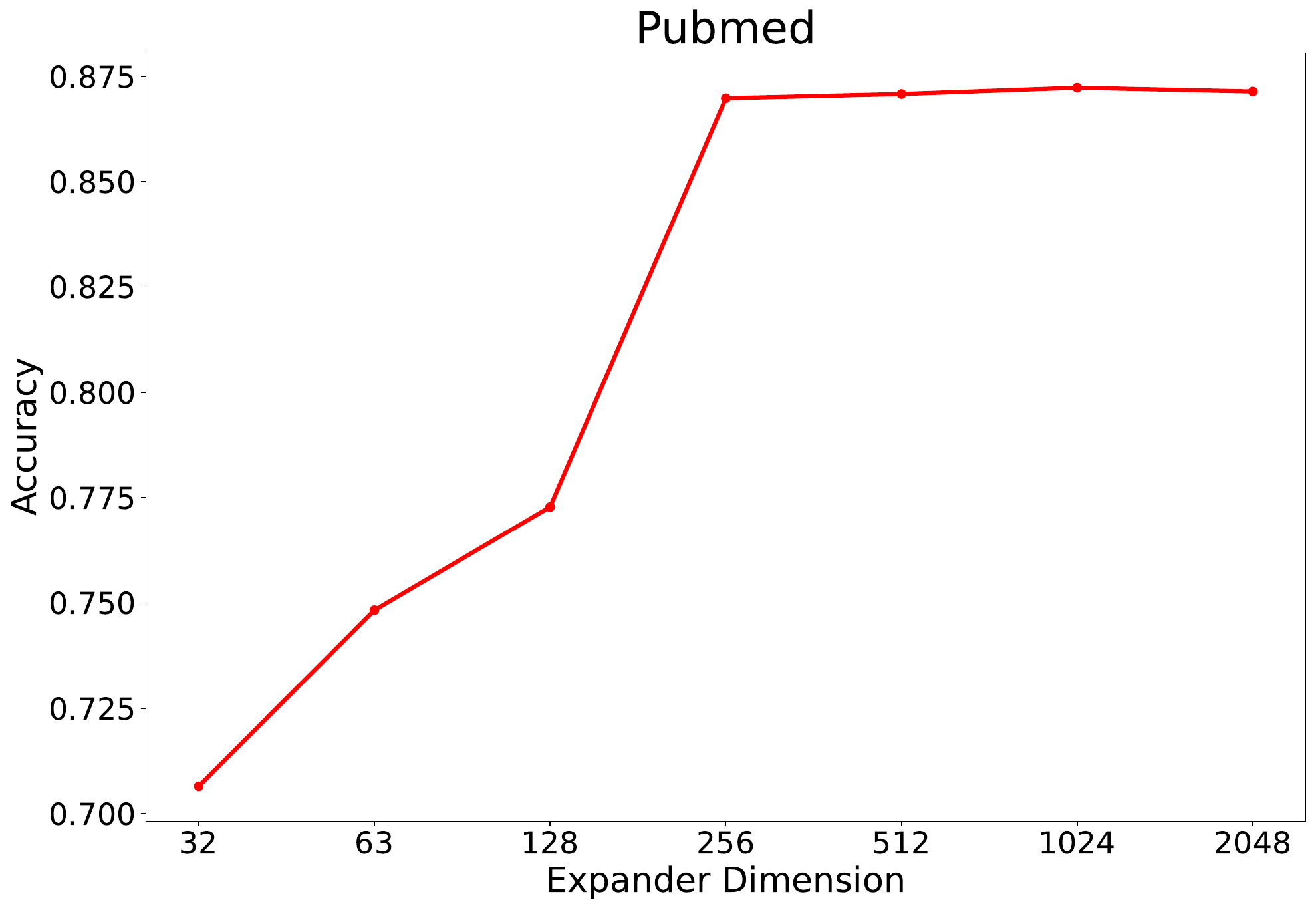}
    \includegraphics[width=0.20\textwidth]{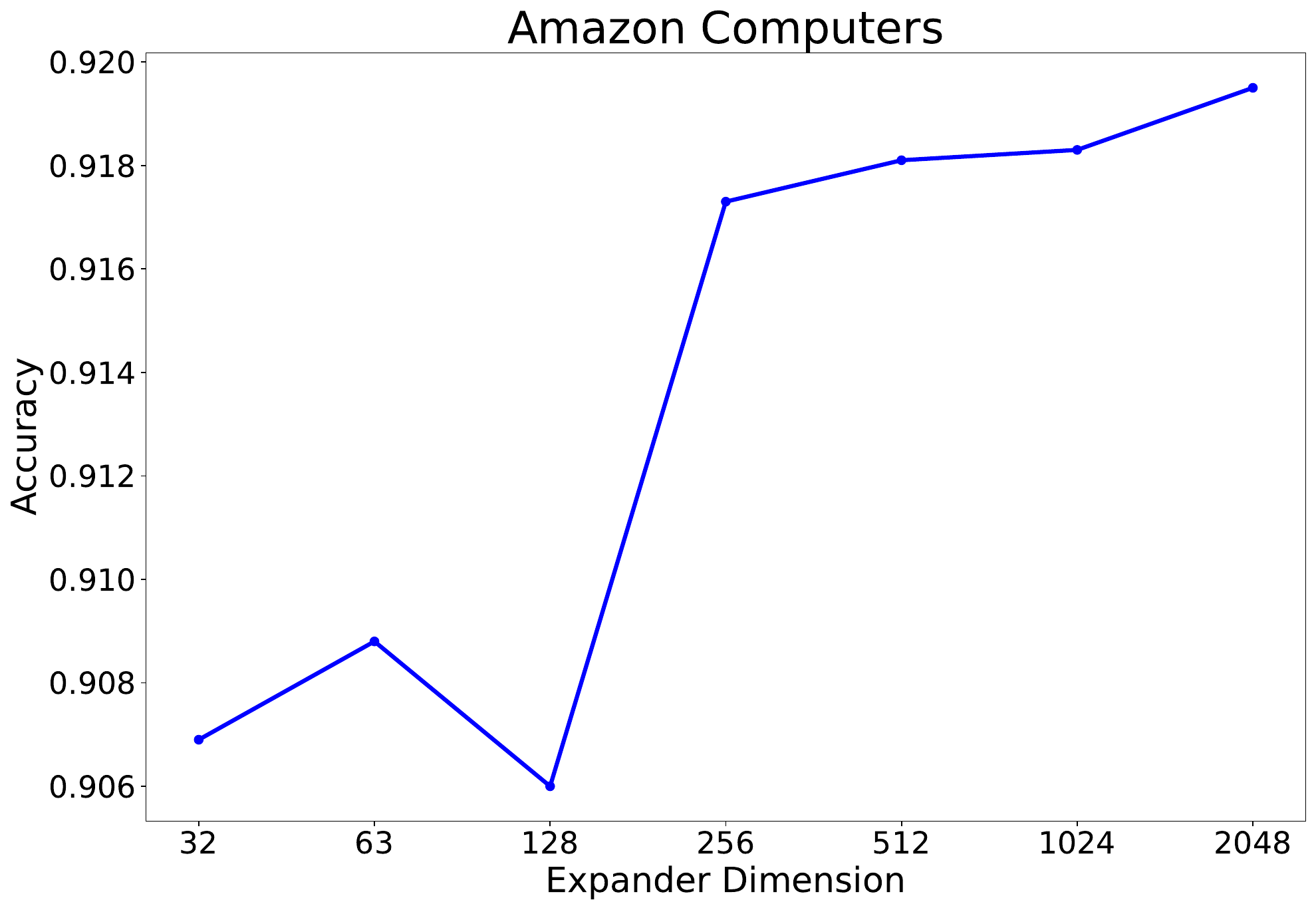}
\caption{Downstream performance of VICReg trained with different expander dimensions. The performance is measured as the mean over 10 trials.}
\label{DimPlot}
\end{figure}

The computational bottleneck in the VICReg loss comes from computing the covariance matrix, which results in an $O(ND^2)$ complexity. As shown in prior work \cite{bardes2022vicreg,zbontar2021barlow,haochen2021provable} on the image domain, both VICReg and Barlow-Twins require the expander $g$ to project the output on a high dimensional $D$. In Figure \ref{DimPlot}, we observe a similar trend in the graph domain across various datasets where increasing embedding dimension impact the downstream performance. To alleviate the computation overhead, we further show in addition to sampling across $N$, dimension sampling can also improve efficiency while maintaining performance.
In this section, we use the Nystrom approximation to refrain from computing and storing the entire covariance matrix and bring down the complexity to $O(NM^2)$, where $M << D$.

\subsection{Nystrom Approximation for Matrix Whitening}
The Nystrom method obtains a low-rank approximation of a positive semi-definite matrix using a subset of its rows and columns. Consider $Cov \in \mathbb{R}^{D \times D}$, the covariance matrix of $Z$, rewritten as block matrix with entries $A \in \mathbb{R}^{M \times M}$, $B \in \mathbb{R}^{M \times (D-M)}$, and $C \in \mathbb{R}^{(D-M) \times (D-M)}$. The variables selected in the block $A$ are commonly known as landmarks. There are various landmark selection methods introduced in the literature \cite{ghojogh2023elements}. We leave a principled investigation of the various landmark selection methods to the future work.

\textit{Nystrom Approximation \cite{williams2000nystrom,ghojogh2023elements}:} for $M$ greater or equal to $rank(Cov)$, the $C$ block in $Cov$ can be accurately approximated as $B^T A^{\dagger} B$:
\begin{equation}
Cov = 
\begin{bmatrix}
  A
  & \vline & B\\
\hline
  B^T & \vline &
  C
\end{bmatrix} =
\begin{bmatrix}
  A
  & \vline & B\\
\hline
  B^T & \vline &
  B^T A^{\dagger} B
\end{bmatrix} =
\begin{bmatrix}
  A
  & \vline & B\\
\hline
  B^T & \vline &
  B^T A^{-1} B
\end{bmatrix} 
\end{equation}
We can replace the pseudo-inverse since $A$ does not have rank deficiency in our particular case. 

\textit{Proof:} Consider a partitioning of $Z =  \begin{bmatrix} z_1 & ...  & z_m & z_{m+1} & ... & z_D \ \end{bmatrix}_{N \times D}$ to matrices R and S, where $R = \begin{bmatrix} z_1 & ...  & z_m \ \end{bmatrix}_{N \times M}$ and $S = \begin{bmatrix} z_{m+1} & ...  & z_D \ \end{bmatrix}_{N \times (D-M)}$. We can write the covariance matrix as:
\begin{equation}
Cov = Z^TZ = \begin{bmatrix} R^T \\ S^T \\ \end{bmatrix} \begin{bmatrix} R^T & S^T \\ \end{bmatrix} = \begin{bmatrix} R^TR & R^TS \\ S^TR & S^TS \end{bmatrix} = \begin{bmatrix}
  A
  & \vline & B\\
\hline
  B^T & \vline &
  C
\end{bmatrix}
\end{equation}
Since $A$ is symmetric, we can write 

$$A = R^TR = U \Sigma U^T$$ 
$$B = R^TS = U \Sigma^{1/2}S$$

Thus we get $S = \Sigma^{-1/2}U^TB$. The approximation in Nystrom can be proved by simply substituting the above for $S$:
$$C = S^TS = B^T U \Sigma^{-1/2}\Sigma^{-1/2}U^TB = B^TA^{-1}B $$ 

In non-contrastive SSL, the goal is to maximize the information by enforcing $Cov$ to be close to the identity matrix. Using Nystrom, we can enforce $A$ and $B^T A^{-1} B$ to be identity. However, inverting $A$ is computationally expensive. Alternatively, we could minimize the norm of $B$ and force $A$ to be identity. However, this would result in $B^T A^{-1} B$ becoming zero. Thus, we resort to making $A$ identity and making $B$ orthonormal. We show that enforcing $A$ to be identity and $B$ orthonormal results in $Cov$ being full rank with equal eigenvalues, which satisfies the VICReg objective of maximizing decorrelation between different variables. Given the covariance matrix $Cov$ in the above block form, if $A = I$ and $B$ is orthonormal, then $Cov = 2.I$:

\begin{equation}
Cov. Cov \approx \begin{bmatrix}
  A
  & \vline & B\\
\hline
  B^T & \vline &
  I
\end{bmatrix} . 
\begin{bmatrix}
  A
  & \vline & B\\
\hline
  B^T & \vline &
  I
\end{bmatrix} =
\begin{bmatrix}
  A^2 + BB^T
  & \vline & AB^T + B^T\\
\hline
  AB + B & \vline &
  B^TB + I
\end{bmatrix} 
=
\begin{bmatrix}
  2I
  & \vline & 2B^T\\
\hline
  2B & \vline &
  2I
\end{bmatrix} =
2 Cov
\end{equation}

Thus, 
\begin{equation}
Cov^2 - 2Cov = 0
\end{equation}
Since we know $Cov \neq 0$, then $Cov$ must equal $2I$. As a result, $Cov$ is orthogonal with equal eigenvalues.

\begin{table*}[!t]
\begin{center}
\resizebox{0.7\pdfpagewidth}{!}{
\begin{tabular}{ccccccc}
\hline        Dataset
                                                & Amz Computers                 & Amz Photos                    & CoauthorCS                    & Coauthor Physics              & DBLP 
                                                        & PubMed\\ \hline
                                                        \midrule
\multicolumn{1}{c|}{VICReg}                & 91.75 $\pm$ 0.42 & 93.79 $\pm$ 0.72 & 94.64 $\pm$ 0.29 & 92.23 $\pm$ 0.27 & 84.36 $\pm$ 0.48 & 86.98 $\pm$ 0.51\\
\multicolumn{1}{c|}{VICReg + NS}      & 91.99 $\pm$ 0.27 & 94.32 $\pm$ 0.57 & 94.02 $\pm$ 0.39 & 95.78 $\pm$ 0.14 &  85.43 $\pm$ 0.41 &  87.12 $\pm$ 0.37 \\
\multicolumn{1}{c|}{VICReg + DS} & 91.82 $\pm$ 0.40 & 94.18 $\pm$ 0.55 & 93.74 $\pm$ 0.29 & 95.66 $\pm$ 0.11 & 85.27 $\pm$ 0.31 & 87.16 $\pm$ 0.50 \\
\multicolumn{1}{c|}{VICReg + JS}        &  91.87 $\pm$ 0.47  &  -  &  -  &  -  &  -  &  87.25 $\pm$ 0.35 \\

\end{tabular}
}
\caption{The downstream accuracy of VICReg SSL approach pre-training the encoding model with different methods of sampling such as Node Sampling (NS), Dimension Sampling (DS), and Joint Sampling (JS). The best results are chosen after the hyper-parameter search over the sampling ratio $p$ and the node sampling methods for each dataset.}
\label{tab:maintable}
\end{center}
\end{table*}

\subsection{Nystrom as Uniform Dimension Sampling}
Enforcing $A$ to be identity and $B$ to be orthonormal requires computing both matrices, resulting in an $O(NM^2) + O(N(D-M)^2)$ computation. This is not significantly better than directly computing the covariance matrix in time $O(ND^2)$. Next, we show that enforcing $A$ to be identity would make  $B$ orthonormal in hindsight. Thus we can omit the extra computation on $B$ and reduce the complexity to $O(NM^2)$. The key to this observation is that $A$ is whitened in every epoch. By sampling different dimensions to form the matrices $A$ and $B$ at each epoch, we can indirectly enforce $B$ to be orthonormal by only enforcing $A$ to be identity. 

\textit{Proposition:} Denote $Cov_i$ the covariance matrix at the end of the $i_{th}$ training epoch, with blocks $A_i$, $B_i$ and $B_i^T A_i^{-1} B_i$. If a different subset of $M$ dimensions are selected to form $A_i$ at each epoch (whitening is applied to a different subset of variables), then $B$ will be an orthonormal matrix at the end of training.

\textit{Proof:} We make two simplifying assumptions; $(i)$ after whitening is applied to block $A_i$ at iteration $i$, the block $A_i$ will remain perfectly whitened ($A_i = I$) $(ii)$ we partition the dimensions into $D/M$ splits and apply whitening to the $i_{th}$ split at iteration $i$. Whitening is applied to every split exactly once.
More concretely, $Cov_i$ is of form

\begin{equation}
Cov_i \approx \begin{bmatrix}
  A_i
  & \vline & B_i\\
\hline
  B_i^T & \vline &
  B_i^T A_i^{-1} B_i
\end{bmatrix} =
\begin{bmatrix}
  I
  & \vline & B_i\\
\hline
  B_i^T & \vline &
  B_i^T B_i
\end{bmatrix} = 
\begin{bmatrix}
  I
  & \vline & R_i^T S_i\\
\hline
  S_i^T R_i & \vline &
  S_i^T S_i
\end{bmatrix}
\end{equation}

Where $R_i = \begin{bmatrix} z_{Mi+1} & ...  & z_{(i+1)M} \ \end{bmatrix}_{N \times M}$ , and $S_i = \begin{bmatrix} z_{1} & ... & z_{Mi} & z_{(i+2)M}  & ... & z_{D} \ \end{bmatrix}_{N \times (D-M)} = \begin{bmatrix} R_1 & .... & R_{i-1} & R_{i+1} & ... & R_M \ \end{bmatrix}_{N \times (D-M)}$ ($S_i$ contains all columns in $Z$ except the columns in $R_i$). 

Note that the $B$ block at the end of training will be of form $B_{M} = R_{M}^T \begin{bmatrix} R_1 & .... & R_{M-1} \ \end{bmatrix}$. Since $A_i$'s have all been whitened in previous epochs, then $A_i = I$ for all $i = 1,...,M-1$. Thus we have $R^T_i. R_i = I$. We can write $B_M$ as:

\begin{equation}
B_M = R_{M}^T \begin{bmatrix} R_1 & .... & R_{M-1} \ \end{bmatrix} =  \begin{bmatrix} R_{M}^T R_1 & .... & R_{M}^T R_{M-1} \ \end{bmatrix} 
\end{equation}

which is orthonormal since $R_{M}^T$ and $R_i$ are each orthonormal and the product of two orthonormal matrices is also orthonormal. Hence, $B$ will be an orthonormal matrix at the of training.

In practice, we sample dimensions uniformly at each epoch to form the block matrix $A_i$. The assumptions in our proof are reasonable since we do this sampling for a long time to cover all the $D$ variables in the covariance matrix.


\section{Experimental Evaluation}
\label{sec5}

\subsection{Experimental Setup}
We follow the standard linear-evaluation protocol on graphs \cite{veličković2018deep,zhu2020deep,thakoor2022largescale}, where the graph encoder is trained in a self-supervised manner. After pre-training is complete, a linear model is trained on top of the frozen embeddings through a logistic regression loss with $l_2$ regularization, while the weight parameters of the encoder is frozen. We consider a GCN  \cite{kipf2017semisupervised} as our encoder, as is the standard in prior works \cite{thakoor2021bootstrapped,zhu2020deep}.

We perform our experiments on a variety graph datasets: $(i)$ classifying predicting article subject categories in citation networks including PubMed and DBLP $(ii)$ product category prediction on Amazon Computers and Photos $(iii)$ and classifying authors as nodes in citation networks (Coauthor CS and Coauthor Physics).

We train VICReg with an expander dimension D on a portion of the N nodes. To ensure low variance across node feature estimations, we feed the entire graph to the GCN encoder. However, the final loss is only computed on the selected sampled nodes or dimensions. For all datasets, we use a two layer expander with the final output dimension $D$ set to 512. For all datasets, we perform a grid search by sampling $p \in \{0.01, 0.1, 0.25, 0.5, 0.75, 1\}$ percent of the nodes at every epoch and computing the loss on the $p \times N$ selected nodes. We perform the same grid search for dimensions sampling on $D$. However, we only compute the covariance loss term on the selected dimensions as it results in slightly higher accuracy compared to when the entire loss term is computed on the selected dimensions. 

\subsection{Experimental Results}
We conduct experiments to show the impact of the sampling methods on the baseline SSL pre-training method, VICReg, measured under the linear probing on the performance of the downstream task. We show the downstream accuracy of the baseline in addition to the three sampling methods: Node Sampling (NS), Dimension Sampling (DS), and Joint Sampling (JS) in Table \ref{tab:maintable}.
The main observation is that for datasets such as Amazon Computers, Amazon Photos, Coauthor Physics, and DBLP, node sampling surpasses not only the baseline but also the other two methods. For PubMed, joint sampling, where a subset of the nodes and dimensions is simultaneously selected,  stands on top of the other methods. Other than Coauthor CS, all sampling methods improve the baseline method. Although the performance improvement is marginal, lower computation complexity on the other hand improves the efficiency significantly. This finding supports the hypothesis that a subset of nodes and dimensions in the embedding space is sufficient for learning representations that are robust for the downstream tasks. This is severe on graph domain where the number of nodes increases across real domains where not all nodes are carrying information with respect to data augmentations. 
We speculate the shortcoming of the sampling methods on Coauthor CS 
may stem from the misalignment in the choices of hyper-parameters of different modules such as the data augmentation, the encoder, and the loss function in VICReg as the regularization coefficients are very sensitive and may simply cause severe dimensional collapse.
We leave a more in-depth investigation on such cases to the future work.

\begin{figure}
     \centering
     \begin{subfigure}[b]{0.4\textwidth}
         \centering
         \includegraphics[width=\textwidth]{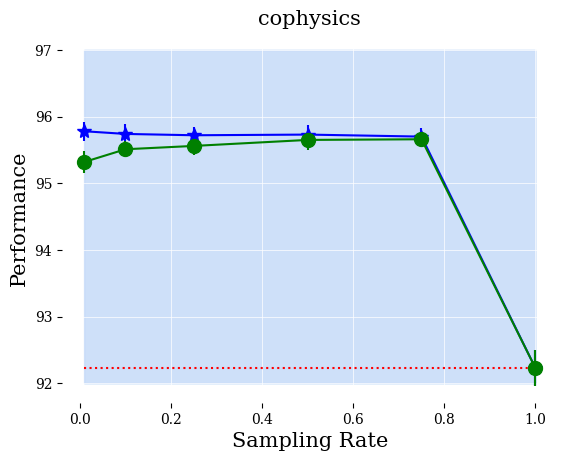}
        \label{fig:subplot1}    
         
     \end{subfigure}
     \begin{subfigure}[b]{0.4\textwidth}
         \centering
         \includegraphics[width=\textwidth]{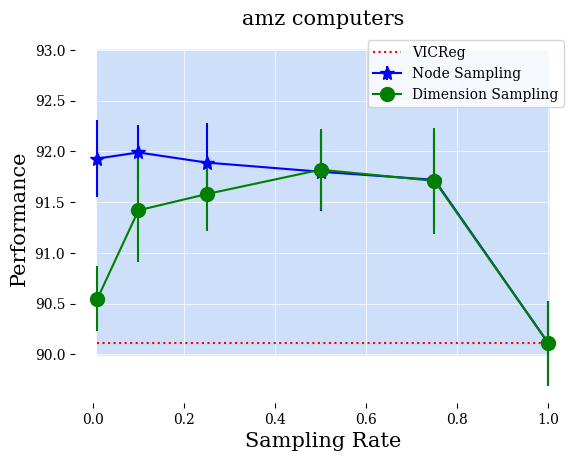}
        \label{fig:subplot2}
     \end{subfigure}
    \caption{Sensitivity Analysis on Sampling Ratio for Dimension and Node Sampling methods. } 
    \label{figure2}
\end{figure}



    
    

\begin{figure}
     \centering
     \begin{subfigure}[b]{0.4\textwidth}
         \centering
         \includegraphics[width=\textwidth]{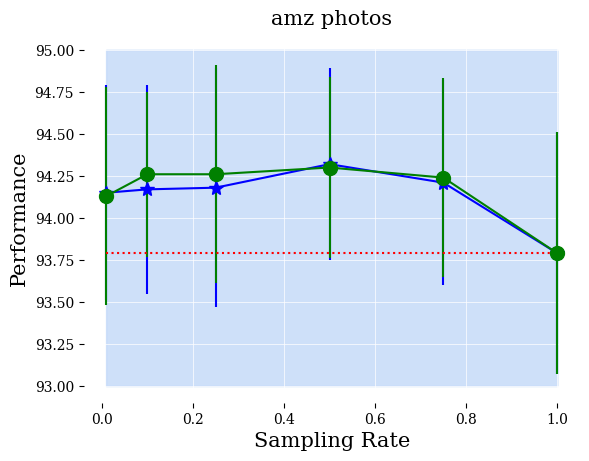}
        \label{fig:subplot3}    
         
     \end{subfigure}
     \begin{subfigure}[b]{0.4\textwidth}
         \centering
         \includegraphics[width=\textwidth]{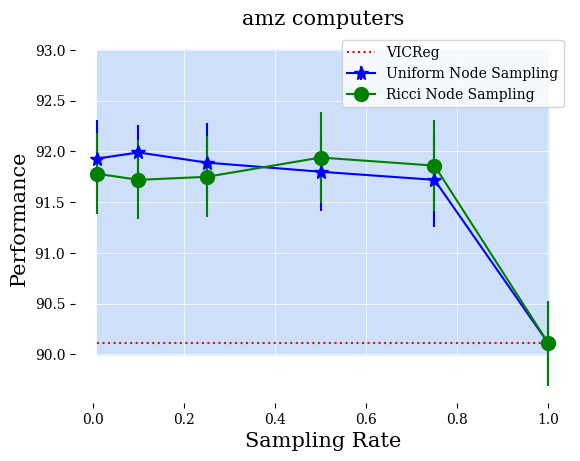}
        \label{fig:subplot4}
     \end{subfigure}
    \caption{Sensitivity Analysis on Sampling Ratio for different Node Sampling methods. } 
    \label{figure3}
\end{figure}




\begin{table}[t]
\small
\centering
\resizebox{0.6\columnwidth}{!}{%
\begin{tabular}{lllll}
\hline
                              & amz computers                      & amz photos                       & cocs                             & cophysics                                                   \\ \hline \midrule
\multicolumn{1}{l|}{0.01 x N} & 91.93 $\pm$0.38   & 94.15 $\pm$0.64 & 94.02 $\pm$0.39 & 95.78 $\pm$0.14  \\
\multicolumn{1}{l|}{0.10 x N} & 91.99 $\pm$0.27   & 94.17 $\pm$0.62 & 93.78 $\pm$0.29 & 95.74 $\pm$0.15  \\
\multicolumn{1}{l|}{0.25 x N} & 91.89 $\pm$0.39   & 94.18 $\pm$0.71 & 93.78 $\pm$0.28 & 95.72 $\pm$0.13  \\
\multicolumn{1}{l|}{0.5 x N}  & 91.80 $\pm$0.39   & 94.32 $\pm$0.57 & 93.80 $\pm$0.26 & 95.73 $\pm$0.15  \\
\multicolumn{1}{l|}{0.75 x N} & 91.72 $\pm$0.46   & 94.21 $\pm$0.61 & 93.74 $\pm$0.32 & 95.70 $\pm$0.13  \\
\multicolumn{1}{l|}{1 x N}    & 91.75 $\pm$ 0.42 & 93.79 $\pm$0.72 & 94.64 $\pm$0.29 & 92.23 $\pm$0.27 
\end{tabular}
}
\caption{VICReg pre-trained with uniform node sampling for various ratio values.}
\label{tab:uniformNode-medium}
\end{table}

\begin{table}[t]
\small
\centering
\resizebox{0.6\columnwidth}{!}{%
\begin{tabular}{lllll}
\hline 
                              & amz computers                    & amz photos                       & cocs                             & cophysics                                                    \\ \hline \midrule
\multicolumn{1}{l|}{0.01 x D} & 90.55 $\pm$0.32 & 93.43 $\pm$0.78 & 93.74 $\pm$0.29 & 95.32 $\pm$0.16  \\
\multicolumn{1}{l|}{0.10 x D} & 91.42 $\pm$0.51 & 93.54 $\pm$0.74 & 93.44 $\pm$0.34 & 95.51 $\pm$0.12 \\
\multicolumn{1}{l|}{0.25 x D} & 91.58 $\pm$0.36 & 93.86 $\pm$0.64 & 93.61 $\pm$0.41 & 95.56 $\pm$0.13  \\
\multicolumn{1}{l|}{0.5 x D}  & 91.82 $\pm$ 0.40 & 94.18 $\pm$0.55 & 93.70 $\pm$0.34 & 95.65 $\pm$0.15  \\
\multicolumn{1}{l|}{0.75 x D} & 91.71 $\pm$0.52 & 94.13 $\pm$0.56 & 93.69 $\pm$0.27 & 95.66 $\pm$0.11  \\
\multicolumn{1}{l|}{1 x D}    & 91.75 $\pm$0.42 & 93.79 $\pm$0.72 & 94.64 $\pm$0.29 & 92.23 $\pm$0.27 
\end{tabular}
}
\caption{VICReg pre-trained with uniform dimension sampling for various ratio values.}
\label{tab:uniformDim-medium}
\end{table}

\subsection{Sensitivity Analysis}
In this section, we provide controlled sensitivity analysis over various design choices such as the sampling ratio and sampling methods. We aim to reveal the underlying factors behind them and pave the path away for the future work to improve the sampling methods.

\textbf{Sampling Ratio:} Given the baseline results, we show how sensitive the downstream linear probing performance is to various levels of sampling in Figure \ref{figure2}. For both Amazon Computers and Coauthor Physics, node sampling results are slightly higher for various ratio values. Furthermore, in Table \ref{tab:uniformNode-medium}, we show results over all sampling ratios for different datasets. It should be noted that for some datasets such as Amazon Computers, the best result is achieved for $0.10$ while for other such as Amazon Photos it is $0.50$. This is likely to be due to the level of semantic information various nodes in each datasets carry. Since some datasets contain more redundant nodes, lower sampling ratio achieves higher performance gain while improving the efficiency of the pre-training stage significantly.
We additionally study the role of dimension sampling ratio in Table \ref{tab:uniformDim-medium}. There is less sensitivity to the ratio for the dimension sampling since the dimension of the embedding space is very critical for a reliable representation learning in the SSL pre-training. According to \cite{haochen2021provable}, the dimension of the embedding space should be at least as high as the number of semantic classes in the data domain. Our empirical finding supports that as the sampling ratio increases, the downstream performance degrades. Furthermore, the accuracy is generally decreased as the model relies on fewer dimensions for loss computation. The trade-off between accuracy and the time-memory cost is interesting; as computing the loss on a few dimensions can make the training procedure much less costly while only affecting the accuracy by a small margin.

\begin{figure}
     \centering
     \begin{subfigure}[b]{0.4\textwidth}
         \centering
         \includegraphics[width=\textwidth]{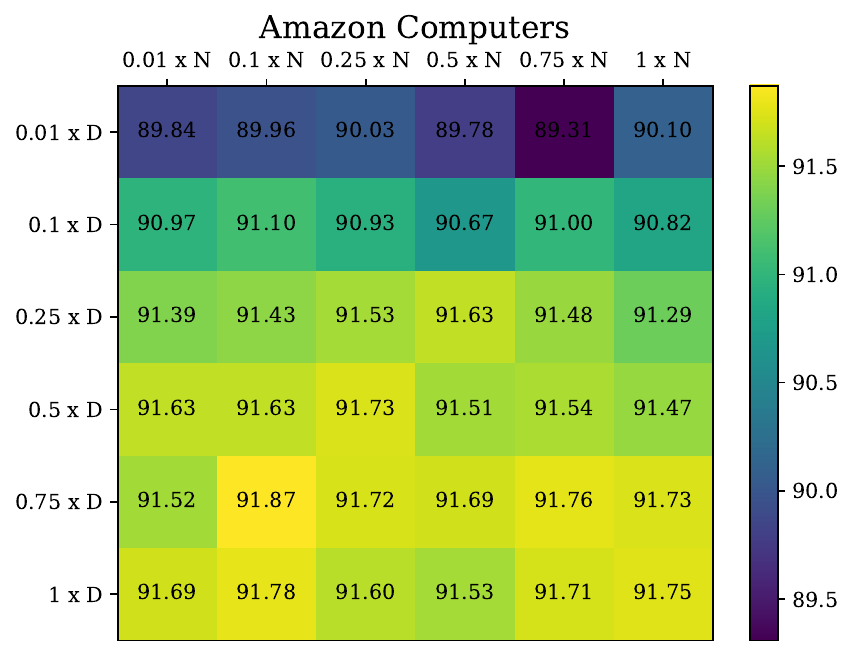}
        \label{fig:subplot5}    
         
     \end{subfigure}
     \begin{subfigure}[b]{0.4\textwidth}
         \centering
         \includegraphics[width=\textwidth]{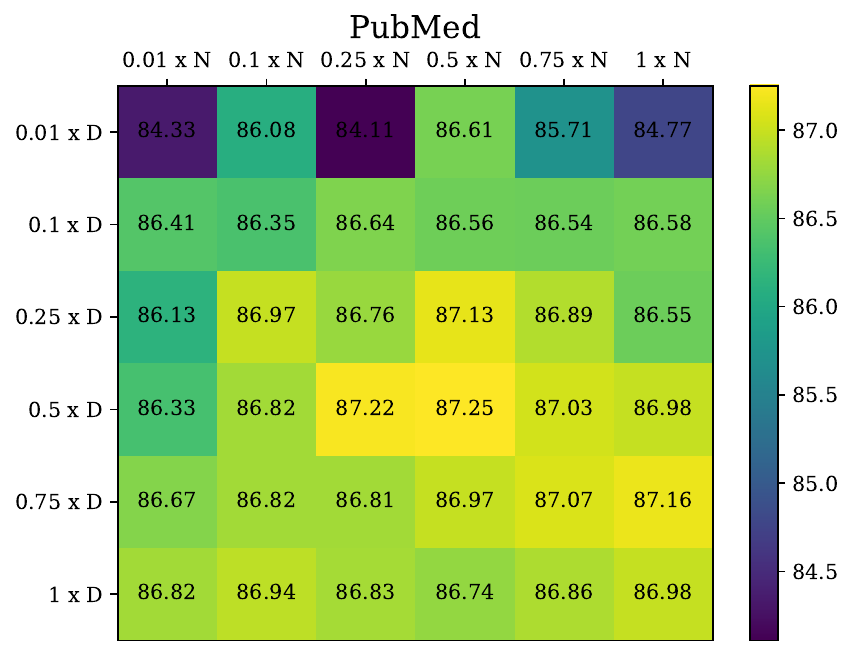}
        \label{fig:subplot6}
     \end{subfigure}
    \caption{The heat-map visualization for the sensitivity analysis on sampling ratio for two datasets: Amazon Computers and PubMed. Random node sampling method is chosen.} 
    \label{fig:jointsamp}
\end{figure}




    

\textbf{Joint Sampling:} To further investigate the correlation between the node and dimension sampling, the results for the grid search over the sampling ratios for joint sampling are presented in Figure \ref{fig:jointsamp}. It is clear that joint sampling empirically outperforms the individual node or dimension sampling for both datasets. In PubMed, more nodes (i.e. \%50) should be included in the pre-training while in Amazon Computers less is optimal (i.e. \%10) while both needs around half of dimension variables in the best case scenario. This supports the argument that the two sampling methods are positively correlated.

\textbf{Node Sampling Methods:} In this work, various node sampling methods can be studied such as uniform, Ricci, and Random Projection (RP). As results presented in Table \ref{tab:various-node-samp}, there is a tie between the Ricci and Uniform sampling. On the small dataset, 
Ricci demonstrates greater success in node sampling, whereas on the larger datasets, Ricci is not as powerful as Uniform sampling. 
We speculate that current benchmark datasets do not contain a lot of bottleneck (low Ricci curvature) nodes. As such, sampling with Ricci does not give us a significant gain in performance. Furthermore, this can be due to the errors in the approximation of the Ricci curvature computation or depending on the complexity of the graph structure in larger graphs, Ricci calculation becomes noisy. We currently calculate Ricci offline in the pre-processing stage given the input node features while it can be calculated online given the intermediate hidden representations. 
We defer the investigation of the latter aspect to the future work.

\begin{table}[h]
\small
\centering
\resizebox{0.6\columnwidth}{!}{%
\begin{tabular}{c|c c c}
\hline
\cline{2-4} 
Node Sampling & PubMed & Amz Comp & CoCS \\
\hline \midrule
Uniform &  86.94 $\pm$ 0.33 & 91.99 $\pm$ 0.27 & 94.02 $\pm$ 0.39 \\
Ricci &  87.12 $\pm$ 0.37 & 91.86 $\pm$ 0.45 & 94.00 $\pm$ 0.36 \\
Random Projection & 85.51 $\pm$ 0.47 & 91.61 $\pm$ 0.35 & 93.45 $\pm$ 0.34 \\
\end{tabular}
}
\caption{The best accuracy results of three different node sampling methods for three datasets: PubMed, Amazon Computer, and Coauthor CS.}
\label{tab:various-node-samp}
\end{table}

In Table \ref{tab:uniformRicci-medium}, we further show the impact of the various sampling ratios using Ricci method for different datasets. 
It is worth mentioning that in certain datasets such as Amazon Computers, optimal results are obtained with a different sampling ratio in Ricci compared to Uniform sampling. However, in the other datasets, it is almost the same. This highlights that some graph datasets inherent difference attributes based on which node sampling is performed.

\begin{table}[h]
\small
\centering
\resizebox{0.6\columnwidth}{!}{%
\begin{tabular}{lllll}
\hline 
                              & amz computers                      & amz photos                       & cocs                             & cophysics                                                   \\ \hline \midrule
\multicolumn{1}{l|}{0.01 x N} & 91.78 $\pm$0.40   & 94.13 $\pm$0.65 & 94.00 $\pm$0.36 & 95.78 $\pm$0.15  \\
\multicolumn{1}{l|}{0.10 x N} & 91.72 $\pm$0.39   & 94.26 $\pm$0.49 & 93.77 $\pm$0.28 & 95.74 $\pm$0.13  \\
\multicolumn{1}{l|}{0.25 x N} & 91.75 $\pm$0.40   & 94.26 $\pm$0.65 & 93.77 $\pm$0.30 & 95.73 $\pm$0.14  \\
\multicolumn{1}{l|}{0.5 x N}  & 91.94 $\pm$0.45   & 94.30 $\pm$0.54 & 93.75 $\pm$0.29 & 95.71 $\pm$0.14  \\
\multicolumn{1}{l|}{0.75 x N} & 91.86 $\pm$0.45   & 94.24 $\pm$0.59 & 93.74 $\pm$0.31 & 95.71 $\pm$0.13  \\
\multicolumn{1}{l|}{1 x N}    & 91.75 $\pm$ 0.42 & 93.79 $\pm$0.72 & 94.64 $\pm$0.29 & 92.23 $\pm$0.27 
\end{tabular}
}
\caption{VICReg pre-trained with Ricci node sampling for various ratio values.}
\label{tab:uniformRicci-medium}
\end{table}

\section{Conclusion}
Graph SSL methods suffer from high computational cost imposed by large-scale graphs projected into high-dimensional embedding space. In this work, We propose node and dimension sampling for reducing the cost of computing the loss function in information maximization non-contrastive SSL methods for graphs. 
We provide mathematical justification on how to formulate the novel idea in a principled manner where it is linked to Nystrom approximation in manifold learning and node pooling in encoding models for graph representation learning. 
Additionally, we present theoretical insights into why dimension sampling would not result in performance degradation. 
We utilize the typical experimental setup for graph SSL methods where the performance of the pre-trained model is measured on the downstream tasks under linear probing. The empirical results over various graph datasets support the hypothesis that not only does the performance of the downstream task marginally improve on average, but also the pre-training efficiency is increased according to the sampling ratio. On both node and dimension sampling side, we plan to deepen our investigation into a more thorough and principled formulation of the sampling methods in the future work. The landmark selection and pooling layers can be further investigated respectively for further in-depth exploration.

\clearpage

\bibliographystyle{unsrt}
\bibliography{ijcai24}

\end{document}